\documentclass[a4paper,11pt,twoside]{article}
\usepackage{mic09}
\usepackage{graphicx}
\usepackage{cite}
\usepackage{url}

\def\id{141} % Please, do not change this field. It will be changed by
\begin{document}

% Title of your final abstract
\mictitle{The Single Machine Total Weighted Tardiness Problem---\\%%
Is it (for Metaheuristics) a Solved Problem ?}

% Your name(s): first name, last name and affiliation mark
\author{
  \micauthor{Martin Josef}{Geiger}{\firstaffiliationmark}
  % \and
  % \micauthor{firstname}{name}{\secondaffiliationmark}
}

% Your affiliation(s), address(es), and e-mail(s) as they should appear
% in the proceedings volume, preceded by the affiliation mark used for the
% respective author(s)
\institutions{
  \micinstitution{\firstaffiliationmark}
  {Logistics Management Department,\\Helmut Schmidt University --- University of the Federal Armed Forces Hamburg}
  {Holstenhofweg 85, 22043 Hamburg, Germany}
  {mjgeiger@hsu-hh.de}
}

\maketitle
\thispagestyle{fancyplain}

\section{Introduction}
The single machine total weighted tardiness problem (SMTWTP) is a
well-known planning problem from operations research, engineering
and computer science. It is characterized by the assignment of
starting times for a given set of jobs on a single processor,
minimizing the (weighted) tardy completion of the jobs with respect
to given due dates. In this sense, the tardiness is given an
economical interpretation, referring to the consideration of costs
which consequently have to be avoided as much as possible.\\Although
most practical problems involve multiple resources (e.\,g.\
machines), many problems can be decomposed into a subsequently
solved series of single-machine problems, as for example done in the
famous shifting bottleneck heuristic \cite{adams:1988:article}. The
effective solution of each single-machine subproblem is therefore a
relevant aspect for solving more complex models.

Besides the problem's relevance with respect to the applicability in
real-world situations, the SMTWTP is computationally challenging, as
it has been proven to be strongly $\mathcal{NP}$-hard
\cite{lenstra:1977:incollection}.\\While some exact methods are
available, many successful solution approaches are based on
heuristics \cite{abdulrazaq:1990:article}. More recently, several
different metaheuristics have been developed for the SMTWTP,
successfully solving benchmark instances from the scientific
literature. Important work includes simple local
search~\cite{maheswaran:2003:article}, Evolutionary
Algorithms~\cite{crauwels:1998:article}, Ant Colony Optimization
\cite{besten:2000:inproceedings,merkle:2000:inproceedings,bauer:1999:inproceedings},
Iterated Local Search \cite{besten:2001:inproceedings}, and
Simulated Annealing~\cite{nearchou:2004:article}. A particularly
successful neighborhood search technique for the SMTWTP is {\em
Iterated Dynasearch}~\cite{congram:2002:article}, which uses dynamic
programming to determine an optimal series of moves to be executed
simultaneously.

While it has been pointed out that available benchmark instances are
comparably easily solvable by local search
\cite{besten:2001:inproceedings}, several recent publications aim to
push the scientific knowledge even further by proposing more refined
metaheuristics
\cite{bilge:2007:article,maheswaran:2005:article,chou:2009:article,panneerselvam:2006:article,wang:2009:article,altunc:2009:article}.
Common to these approaches is the fact that they lead to rather
impressive results within the chosen experimental settings. From
this perspective, and taking into account the remarks of
\cite{besten:2001:inproceedings}, we need to raise the question
whether recent results are so surprising after all. Or is it
possible that, despite the problem's complexity, the SMTWTP is a
`solved' problem for metaheuristics? Are there `difficult'
instances, and if so, how can they be described in terms of their
technical properties? Can we draw conclusions on how to organize
future research?

The article is organized as follows. Section~\ref{dsadsa:problem}
presents a quantitative model for the SMTWTP and describes an
appropriate representation (coding) for the alternatives. Besides,
benchmark instances from the literature are briefly introduced. The
following Sections~\ref{dsadsa:hillclimbing} and~\ref{dsadsa:vns}
revisit local search for the SMTWTP, namely hillclimbing and
Variable Neighborhood Search \cite{hansen:2003:incollection}. The
identification of comparable `difficult' instances is derived from
the experiments, and a closer analysis of these data sets follows in
Section~\ref{dsadsa:difficult:instances}. Conclusions are presented
in Section~\ref{dsadsa:conclusions}.

\section{\label{dsadsa:problem}Problem statement and benchmark data}
\subsection{Quantitative model}
In the SMTWTP, a set of jobs $\mathcal{J} = \{ J_{1}, \ldots, J_{n}
\}$ needs to be processed on a single machine. Each job $J_{j}$
consists of a single operation only, involving a processing time
$p_{j} > 0 \,\,\forall j = 1, \ldots, n$. The relative importance of
the jobs is expressed via a nonnegative weight $w_{j}
> 0 \,\, \forall j = 1, \ldots, n$. Processing on the machine is only possible for a single job at a
time, excluding parallel processing of jobs. Each job $J_{j}$ is
supposed to be finished before its due date $D_{j}$. If this is not
the case, a tardiness $T_{j}$ occurs, measured as $T_{j} = \max
\left\{ s_{j} + p_{j} - D_{j} ; 0 \right\}$, where $s_{j}$ denotes
the starting time of job $j$. The overall objective of the problem
is to find a feasible schedule $x$ minimizing the total weighted
tardiness $TWT$, i.\,e. $\min\,\, TWT = \sum_{j = 1}^{n} w_{j}
T_{j}$.

A schedule for a particular problem can be interpreted as a vector
of starting times of the jobs, $x = ( s_{1}, \ldots, s_{n} )$. We
assume that processing starts at time 0, thus $s_{j} \geq 0 \,\,
\forall j = 1, \ldots, n$. A possible overlapping of jobs on the
machine is avoided by the formulation of disjunctive side
constraints: $s_{j} \geq s_{k} + p_{k} \,\, \dot\vee \,\, s_{j} +
p_{j} \leq s_{k} \,\, \forall j, k = 1, \ldots, n, j \neq k$.

As the objective function of the single machine total weighted
tardiness problem is a {\em regular}
function~\cite{baker:1974:book}, it is known that at least one
active schedule exists which is also optimal. A schedule is called
{\em active} if, for a given sequence of jobs, all operations are
started as early as possible, thus avoiding all unnecessary
in-between waiting times (delays). The problem of finding an optimal
schedule may therefore be reduced to the problem of finding an
optimal sequence of jobs. A given sequence is represented by a
permutation $\pi = \{ \pi_{1}, \ldots, \pi_{n} \}$ of the job
indices. Each element $\pi_{i}$ in $\pi$ stores the index of the job
which is to be processed as the $i$th job in the processing
sequence. The permutation of indices is then \lq{}decoded\rq{} into
a schedule by assigning $s_{\pi_{1}} = 0$ and computing the values
of $s_{\pi_{i}} = s_{\pi_{i-1}} + p_{\pi_{i-1}} \,\,\forall i = 2,
\ldots, n$.\\Obviously, this leads to an active schedule without any
waiting times between jobs.

\subsection{Benchmark instances from the literature}
Optimization approaches for the SMTWTP are commonly verified using
the benchmark instances of \cite{crauwels:1998:article}. The authors
presented 375 data sets with varying characteristics. For values of
$n = 40$, $n = 50$, and $n = 100$, 125 instances are each proposed.
The computation of the processing times $p_{j}$ is randomly drawn
from a uniform distribution $[1,100]$, while the weights are taken
from $[1,10]$. Depending on the relative range of due dates $RDD$
and the average tardiness factor $TF$, the due dates are randomly
computed as integer values within $\left[ P \left( 1 - TF -
\frac{RDD}{2} \right), P \left(1 - TF + \frac{RDD}{2} \right)
\right]$, where $P = \sum_{j = 1}^{n}p_{j}$. Five instances have
been computed for each combination of $RDD$ and $TF$: $RDD \in
\left\{ 0.2, 0.4, 0.6, 0.8, 1.0 \right\}$, $TF \in \left\{ 0.2, 0.4,
0.6, 0.8, 1.0 \right\}$.

All data sets are available in the OR-Library under
\url{http://people.brunel.ac.uk/~mastjjb/jeb/info.html}. For the
ones with $n = 40$ and $n = 50$, optimal solutions are known, while
for the ones of size $n = 100$, at least to our knowledge, only
best-known solutions are reported in the literature, lacking their
final proof of optimality. It should be noticed however, that the
results for the larger data sets are commonly assumed to be optimal,
as, despite rather active research in this area, there has not been
any improvement of the best-known upper bounds within past ten
years.

\section{\label{dsadsa:hillclimbing}Hillclimbing revisited}
\subsection{Configuration}
Despite the existence of advanced metaheuristics for the SMTWTP, we
first investigate the effectiveness of simple local search in a
hillclimbing framework. In these experiments, we are interested in
finding out how good results may get when only applying rather basic
search strategies. In conclusions, such experiments present a lowest
possible benchmark for any alternative technique, and allow an
insight into the difficulty of different benchmark data sets.

The implemented hillclimbing algorithm is based on a
best-improvement move strategy, investigating all neighboring
alternatives and carrying out a move that realizes the largest
possible improvement. Applied neighborhood operators are exchange
({\tt EX}), forward shift ({\tt FSH}) and backward shift ({\tt
BSH}). {\tt EX} exchanges two jobs at the positions $i$ and $j$ in
$\pi$, {\tt FSH} moves a job from position $i$ to $j$ with $i < j$,
and {\tt BSH} moves a job from $j$ to $i$, $i < j$. 100 test runs
have been carried out for each data set/configuration of the
algorithm, each starting from a (different) random job permutation.
Naturally, the test runs terminate with the identification of a
local optimum.

\subsection{Results}
Table~\ref{tbl:number:optimal} shows the number of instances which
have been solved to optimality in at least one of the 100 test runs,
depending on the chosen neighborhood operator and the size of the
instances $n$.

\begin{table}[!ht]
\begin{center}
\caption{\label{tbl:number:optimal}Number of instances for which
optimal/best known solutions have been found by hillclimbing}
\begin{tabular}{lrrr}
\hline%
Operator & $n = 40$ & $n = 50$ & $n = 100$\\%
\hline%
{\tt    EX} & 108 & 85 & 48\\
{\tt   FSH} &  86 & 65 & 31\\
{\tt   BSH} &  48 & 35 & 13\\
\hline
\end{tabular}
\end{center}
\end{table}

It is interesting to see that the results differ rather drastically
with respect to the chosen neighborhood structure. {\tt EX} clearly
leads to the best results, followed by {\tt FSH} and {\tt BSH}. As
it has to be expected, the number of data sets that have been solved
to optimality decreases with increasing $n$. Nevertheless, a large
portion of the 125 instances is solved for each value of $n$, even
in case of the largest value $n = 100$. It becomes apparent that the
benchmark instances given in \cite{crauwels:1998:article} contain a
number of data sets that are less appropriate for state-of-the-art
metaheuristics, as they are comparably easily solvable using {\tt
EX}-hillclimbing.

\begin{table}[!ht]
\begin{center}
\caption{\label{tbl:optimal:wrt:rdd:tf}Number of instances solved to
optimality by hillclimbing depending on $RDD$ and $TF$}
\begin{tabular}{lr|rrrrr|rrrrr|rrrrr}
\hline
&& \multicolumn{5}{c|}{\rule[0mm]{0mm}{5mm} {\tt EX}} & \multicolumn{5}{c|}{{\tt FSH}} & \multicolumn{5}{c}{{\tt BSH}}\\
&      $TF$:& 0.2 & 0.4 & 0.6 & 0.8 & 1.0 & 0.2 & 0.4 & 0.6 & 0.8 & 1.0 & 0.2 & 0.4 & 0.6 & 0.8 & 1.0\\
\cline{2-17}
$RDD$:& 0.2 &  11 &  13 &   7 &  10 &  15 &   5 &   6 &   3 &   7 &  13 &   0 &   0 &   0 &   2 &  12\\
      & 0.4 &  11 &   9 &   6 &  11 &  15 &  12 &   8 &   8 &   6 &   9 &   3 &   0 &   0 &   4 &   9\\
      & 0.6 &  15 &   6 &   3 &   9 &  11 &  15 &   4 &   4 &   3 &   9 &  12 &   0 &   0 &   5 &   6\\
      & 0.8 &  15 &   4 &   4 &   7 &  12 &  15 &   6 &   5 &   2 &   6 &  13 &   0 &   0 &   1 &   7\\
      & 1.0 &  15 &   8 &   5 &   8 &  11 &  15 &  11 &   3 &   1 &   6 &  13 &   4 &   0 &   1 &   4\\
\hline
\end{tabular}
\end{center}
\end{table}

A closer analysis indicates possible characteristics of relatively
`hard' versus `easy' instances. Table~\ref{tbl:optimal:wrt:rdd:tf}
shows the number of instances solved to optimality depending on the
values of $RDD$ and $TF$. As we here aggregate over all settings of
$n$, the maximum possible value in each cell is 15.\\It is possible
to see that instances generated by assuming either low or high
values of $TF$, i.\,e.\ $TF \in \left\{ 0.2, 1.0 \right\}$, turn out
to be comparably easily solvable using simple local search. With
respect to the range of due dates $RDD$, the tendency towards
difficult versus easy data sets is less obvious. In combination with
values of $TF = 0.2$, high values of $RDD$ such as $RDD \in \left\{
0.6, 0.8, 1.0 \right\}$ lead to easy instances. For a value of $TF =
1.0$, the choice of a low $RDD$, i.\,e.\ $RDD \in \left\{ 0.2, 0.4
\right\}$, results in easier data sets.\\On the other hand, values
such as $TF = 0.6$ show the tendency to produce relatively difficult
instances, at least for simple local search based on hillclimbing.

\section{\label{dsadsa:vns}Variable neighborhood search}
\subsection{Configuration}
On the basis of past studies of local search for the SMTWTP
\cite{besten:2001:inproceedings}, and following the outcomes of the
previous section, we investigate two different configurations of
Variable Neighborhood Search (VNS). The first, {\tt VNS-1}, applies
the basic neighborhood operators in order of {\tt
EX}$\rightarrow${\tt FSH}$\rightarrow${\tt BSH}, while the second,
{\tt VNS-2}, implements the reverse order {\tt BSH}$\rightarrow${\tt
FSH}$\rightarrow${\tt EX}. Knowing that there appears to be a
relative order of neighborhood operators, it will be interesting to
further investigate the effects of the different configurations of
the VNS algorithms.

All neighborhoods are searched in a best-move-fashion, thus
searching the entire neighborhood. As usual in VNS, the neighborhood
is switched to the succeeding operator if the active one fails to
improve the current solution. Search terminates with the
identification of a local optimum, which is, in case of VNS, an
alternative being locally optimal with respect to all considered
neighborhoods.\\Again, 100 test runs have been carried out for each
benchmark instance/VNS-configuration, each starting from a
(different) random job permutation.

\subsection{Results}
Table~\ref{tbl:number:optimal:VNS} shows the number of instances
which have successfully been solved by the VNS algorithms in at
least a single out of the 100 test runs.

\begin{table}[!ht]
\begin{center}
\caption{\label{tbl:number:optimal:VNS}Number of instances for which
optimal/best known solutions have been found by VNS}
\begin{tabular}{lrrr}
\hline%
Algorithm & $n = 40$ & $n = 50$ & $n = 100$\\%
\hline%
{\tt VNS-1} ({\tt
EX}$\rightarrow${\tt FSH}$\rightarrow${\tt BSH}) & 121 & 114 & 108\\
{\tt VNS-2} ({\tt BSH}$\rightarrow${\tt
FSH}$\rightarrow${\tt EX}) & 125 & 124 & 117\\
\hline
\end{tabular}
\end{center}
\end{table}

We can see that {\tt VNS-2} is able to solve more instances than
{\tt VNS-1}. It can be suspected that this behavior originates from
the relative order of neighborhood operators. {\tt VNS-2} starts
with the weak operator {\tt BSH}, and then continues to relatively
better neighborhoods. Therefore, an improvement of solutions being
locally optimal to the initially searched neighborhoods appears to
be more likely as in the case of {\tt VNS-1}, where search is
organized starting with relatively good operators.\\Again, the
number of solved instances is decreasing with increasing $n$.
However, the results remain on a rather high level despite the
growing difficulty of the benchmark instances. Only few instances
remain unsolved.

\begin{table}[!ht]
\begin{center}
\caption{\label{tbl:optimal:wrt:rdd:tf:VNS}Number of instances solved to
optimality by VNS depending on $RDD$ and $TF$}
\begin{tabular}{lr|rrrrr|rrrrr}
\hline
&& \multicolumn{5}{c|}{\rule[0mm]{0mm}{5mm} {\tt VNS-1}} & \multicolumn{5}{c}{{\tt VNS-2}} \\
&      $TF$:& 0.2 & 0.4 & 0.6 & 0.8 & 1.0 & 0.2 & 0.4 & 0.6 & 0.8 & 1.0 \\
\cline{2-12}%%
$RDD$:& 0.2 & 15 & 14 & 11 & 13 & 15 & 15 & 15 & 15 & 15 & 15\\
      & 0.4 & 13 & 14 & 14 & 12 & 15 & 15 & 15 & 14 & 14 & 15\\
      & 0.6 & 15 & 11 & 11 & 15 & 15 & 15 & 15 & 15 & 15 & 14\\
      & 0.8 & 15 & 11 & 13 & 15 & 15 & 15 & 15 & 12 & 14 & 15\\
      & 1.0 & 15 & 14 & 12 & 15 & 15 & 15 & 14 & 15 & 14 & 15\\
\hline
\end{tabular}
\end{center}
\end{table}

More detailed results are given in
Table~\ref{tbl:optimal:wrt:rdd:tf:VNS}, which shows the number of
solved instances with respect to the choices of $RDD$ and $TF$. In
comparison to the results obtained by hillclimbing, a similar
pattern arises in the investigation of VNS. Again, both small and
high values of $TF$ lead to easy instances. In the light of the
overall impressive results of VNS however, this pattern is less
obvious. Knowing that {\tt VNS-2} successfully solved 97.6\% of the
benchmark instances, only few instances are left that can be
considered to be `difficult'.

Besides the best-case behavior of the algorithms, the average
quality of the local optima over all 100 test runs is interesting.
Table~\ref{tbl:average:deviation:VNS} gives the results with respect
to this.

\begin{table}[!ht]
\begin{center}
\caption{\label{tbl:average:deviation:VNS}Average deviation from the
optimal/best known solution}
\begin{tabular}{lrrr}
\hline%
Algorithm & $n = 40$ & $n = 50$ & $n = 100$\\%
\hline%
{\tt VNS-1} ({\tt
EX}$\rightarrow${\tt FSH}$\rightarrow${\tt BSH}) & 2.59\% & 2.36\% & 2.10\%\\
{\tt VNS-2} ({\tt BSH}$\rightarrow${\tt
FSH}$\rightarrow${\tt EX}) & 0.99\% & 1.45\% & 0.98\%\\
\hline
\end{tabular}
\end{center}
\end{table}

Also on an average level, {\tt VNS-2} leads to better results than
{\tt VNS-1}. Moreover, the average deviation from the optimal/best
known solutions is rather small, which indicates that VNS is indeed
an effective approach for most instances from the literature.

Unfortunately however, a considerable dispersion can be identified
within these results. As shown in
Table~\ref{tbl:average:deviation:vns:wrt:rdd:tf}, the average
results deviate for some few instances significantly more than in
the majority of data sets. Even for some smaller instances, i.\,e.\
$n = 40$ and $n = 50$, considerable deviations can be found in some
cases.

\begin{table}[!ht]
\begin{center}
\caption{\label{tbl:average:deviation:vns:wrt:rdd:tf}Average
deviation of {\tt VNS-2} from the optimal/best known solutions given
in percent}
\begin{tabular}{lr|rrrrr|rrrrr|rrrrr}
\hline
&& \multicolumn{5}{c|}{\rule[0mm]{0mm}{5mm}$n = 40$} & \multicolumn{5}{c|}{$n = 50$} & \multicolumn{5}{c}{$n = 100$}\\
&      $TF$:& 0.2 & 0.4 & 0.6 & 0.8 & 1.0 & 0.2 & 0.4 & 0.6 & 0.8 & 1.0 & 0.2 & 0.4 & 0.6 & 0.8 & 1.0\\
\cline{2-17}
$RDD$:& 0.2 & 6.7 & 0.4 & 0.2 & 0.2 & 0.0 & 1.5 & 0.4 & 0.4 & 0.1 & 0.0 & 0.4 & 0.2 & 0.2 & 0.2 & 0.0\\
      & 0.4 & 0.0 & 0.3 & 0.6 & 0.1 & 0.0 & 13.9 & 1.2 & 0.4 & 0.1 & 0.0 & 1.1 & 1.1 & 0.4 & 0.1 & 0.0\\
      & 0.6 & 0.0 & 4.6 & 0.9 & 0.0 & 0.0 & 0.0 & 3.1 & 1.8 & 0.2 & 0.0 & 0.0 & 2.5 & 0.6 & 0.1 & 0.0\\%
      & 0.8 & 0.0 & 9.7 & 0.3 & 0.0 & 0.0 & 0.0 & 10.3 & 0.8 & 0.2 & 0.0 & 0.0 & 15.9 & 1.1 & 0.1 & 0.0\\%
      & 1.0 & 0.0 & 0.0 & 0.6 & 0.1 & 0.0 & 0.0 & 0.0 & 1.9 & 0.1 & 0.1 & 0.0 & 0.2 & 0.3 & 0.1 & 0.0\\%
\hline
\end{tabular}
\end{center}
\end{table}

Combining the results of Table~\ref{tbl:average:deviation:VNS} and
\ref{tbl:average:deviation:vns:wrt:rdd:tf}, an interesting
observation emerges. While {\tt VNS-2} successfully solves the
instances with $TF=0.2$/$RDD=0.2$, $TF=0.4$/$RDD=0.6$,
$TF=0.4$/$RDD=0.8$ in the {\em best case}, the results show a
relative high deviation from the optimal/best known results in the
{\em average case}. On the other hand, instances that are not solved
in the best case, i.\,e.\ the ones with $TF=0.6$/$RDD=0.8$, are
approximated very nicely in average.

\subsection{Computational effort}
The following Table~\ref{tbl:evaluations} gives an overview about
the required computational effort for finding a local optimum,
depending on the neighborhood operator/ metaheuristic and the
problem size $n$.

\begin{table}[!ht]
\begin{center}
\caption{\label{tbl:evaluations}Average number of evaluations for finding a local optimum}
\begin{tabular}{lrrr}
\hline
& $n=40$ & $n= 50$ & $n=100$\\
\hline
{\tt EX}   & 23,147 &  46,355 &  416,920\\
{\tt FSH}  & 24,866 &  51,471 &  513,764\\
{\tt BSH}  & 29,563 &  62,637 &  693,206\\
{\tt VNS-1}& 26,389 &  52,692 &  471,406\\
{\tt VNS-2}& 39,847 &  85,420 &  950,904\\
\hline
\end{tabular}
\end{center}
\end{table}

Clearly, the length of the hillclimbing runs significantly increases
with $n$, and it becomes clear that for large values of $n$, the
identification of even a local optimum becomes intractable. On the
other hand, the basic operators {\tt EX}, {\tt FSH} and {\tt BSH} do
not show bigger differences. On the contrary, the relatively
best-performing neighborhood {\tt EX} requires less computations for
finding (better) local optima. Consequently, the effort for the
execution of {\tt VNS-1} is much smaller in comparison to {\tt
VNS-2}, as the latter implementation first invests in the
computationally demanding identification of qualitatively inferior
local optima for {\tt BSH}.

\section{\label{dsadsa:difficult:instances}A first analysis of `difficult' instances}
While the results of the experiments above clearly show that already
simple local search strategies are rather effective for the SMTWTP,
some few instances remain which can be considered to be relatively
`difficult'. In fact, one of the data sets with $n = 50$ and three
with $n = 100$ have not been solved in at least a single run of any
local search strategy. For some other instances with $n = 100$, an
optimal solution has been found, however only in a single test run.
The following Table~\ref{tbl:difficult:instances} gives an overview
about these instances and some of their properties.

\begin{table}[!ht]
\begin{center}
\caption{\label{tbl:difficult:instances}`Difficult' instances and
some of their properties. Unsolved instances are highlighted $\star$.}
\begin{tabular}{lrrrrrr}
\hline%
Instance &   $n$ &  $RDD$ & $TF$ & Distinct optimal alt.  &  Entropy & Entropy of 100 alt.\\%
\hline
50\_109 $\star$ & 50 & 1.0 &  0.4 &$>$ 1,000,000 & 9.42E$-$11  &  0.0036\\
100\_20 & 100 &0.2 &0.8 &$\geq$ 2,304  &  4.52E$-$06 &   0.0018\\
100\_42 & 100 &0.4 &0.8 &$\geq$ 23,552 &  9.60E$-$08 &   0.0026\\
100\_44 & 100 &0.4 &0.8 &$\geq$ 17,280 &  1.24E$-$07 &   0.0021\\
100\_45 $\star$ & 100 &0.4 &0.8 &$\geq$ 52,992 &  1.92E$-$08 &   0.0028\\
100\_86 $\star$ & 100 &0.8 &0.6 &$>$ 100,000&  6.10E$-$09 &   0.0027\\
100\_88 $\star$ & 100 &0.8 &0.6 &$>$ 100,000&  6.31E$-$09 &   0.0028\\
100\_111& 100 &1.0   &0.6 &$\geq$ 3,480  &  1.81E$-$06 &   0.0018\\
100\_113& 100 &1.0   &0.6 &$>$ 100,000&  9.17E$-$09 &   0.0037\\
100\_118& 100 &1.0   &0.8 &$\geq$ 256 & 5.51E$-$04  &  0.0033\\
\hline
\end{tabular}
\end{center}
\end{table}

Obviously, difficult instances are generated by assuming medium
values of $TF$, i.\,e.\ mainly $TF = 0.6$ and $TF = 0.8$. With
respect to the choice of the $RDD$, the tendency towards difficult
data sets is less obvious.

After having been able to identify optimal alternatives even for the
instances that remained unsolved by VNS and hillclimbing, we tried
to estimate the number of distinct global optima for each of the
`difficult' data sets. A large number of distinct optimal
alternatives has been found. In some cases even, the computation of
distinct global optima has been externally terminated after
identifying 1,000,000 (the case of $n = 50$) or 100,000 (the cases
of $n = 100$) alternatives.\\Despite the fact that the search space
comprises $n!$ alternatives, these numbers are surprisingly high.
Although identifying an optimal alternative still is a complex issue
in general, we are able to show that there is a rather large number
of distinct global optima.

Besides the sheer amount of distinct global optima, we have to
question whether these alternatives share similar characteristics,
or whether they are rather different. One possibility for such an
analysis is based on measuring the entropy of the codings as
described in \cite{mattfeld:1999:article}. In this approach, we
measure the distribution of precedence relations between jobs in a
finite set of solutions. Possible values range from 1 for an
entirely random population to 0 for identical codings. As stated in
\cite{mattfeld:1999:article}, let $\omega_{ijk}$ denote the observed
number of solutions with $\textrm{prec}_{i}(j,k) = \mathrm{true}$ in
a pool of size $\mu$. The entropy $E_{ijk}$ of this precedence
relation of operations is given as $E_{ijk} = \frac{-1}{\log
\sqrt{2}}\frac{\omega_{ijk}}{\mu}\log \frac{\omega_{ijk}}{\mu}$, and
the overall entropy is computed as the average over all $E_{ijk}, (j
\neq k)$.

As shown in Table~\ref{tbl:difficult:instances}, the entropy of the
set of optimal alternatives turns out to be extremely small. The
obtained values arise from two observations. First, the codings of
the alternatives are indeed rather similar. Second, the huge number
of distinct global optima leads to the computation of, in average,
such small entropies. As the latter circumstance biases the analysis
of the entropy, we also computed the entropy of a subset of 100
randomly chosen distinct global optima, given in the rightmost
column of Table~\ref{tbl:difficult:instances}. Again, the observed
values are very small, demonstrating that in fact the optimal
alternatives, while generally being distinct, really are
`micro-mutations' of each other.

\section{\label{dsadsa:conclusions}Summary and conclusions}
The current article presented an investigations of some local search
heuristics for the the single machine total weighted tardiness
problem. In particular, we considered several issues raised in
previous work, indicating that existing benchmark instances are not
sufficiently challenging for modern metaheuristics. Therefore, our
experiments first employed a rather simple hillclimbing
algorithm.\\In these first experiments, we have been able to
identify an order of neighborhood operators with respect to their
relative performance. For the SMTWTP, the exchange neighborhood
operator leads to significantly better results than shift
operators.\\Despite the problem's complexity, many data sets have
successfully been solved by simple hillclimbing. We also have been
able to show that the control parameters used for the generation of
the data sets have a significant influence on the difficulty of the
benchmark instances.

Extended experiments have been carried out using Variable
Neighborhood Search. On the basis of the experiments with the
hillclimbing algorithms, an order of neighborhood operators within
the VNS metaheuristics has been tested. The obtained results are
indeed impressive. Even for the larger-sized instances, only few
cases remain unsolved, and the average deviations to the
optimal/best known solutions are surprisingly small.

A first investigation of data sets that, during the experiments,
turned out to be `difficult' followed. We have been able to show
that a large number of distinct global optima exists for these
instances. To some extent we may reason that the sheer number of
global optima makes the instances attractive for local search.

Several conclusions arise from the experiments. While we cannot
generally state that the SMTWTP is a solved problem as such,
existing benchmark instances indeed appear to be unsuitable for the
verification of novel metaheuristics. In this light, recent results
are not too surprising, as even rather simple search strategies will
lead to very close approximations of the optimal solutions of the
established data sets. Before presenting novel metaheuristics for
the SMTWTP and testing them on existing data sets, future research
should therefore address the following open issues first:

\begin{enumerate}
\item {\em Proposition of challenging benchmark instances}.\\Clearly,
bigger data sets are needed, significantly increasing the set of $n$
jobs from $n = 100$ to a larger number. This would prevent the
application of simple hillclimbing/VNS algorithms as the computation
of a locally optimal alternative becomes (prohibitively) costly.
When designing such instances, values of $TF \in \left\{ 0.6, 0.8
\right\}$ should be favored, while $TF \in \left\{ 0.2, 1.0
\right\}$ should be avoided.

\item {\em Theoretical foundation/explanation of the obtained
results}.\\While the observation made with respect to the different
behavior of the neighborhood operators is interesting, there appears
to be a lack of theoretical foundation of these results. We believe
it would be beneficial to explain the obtained results from a
theoretical point of view, as opposed to a purely
experimentally-driven research.

A first attempt in this direction could investigate the relative
impact of the neighborhoods on the derived neighboring solutions.
Based on such an investigation, it might be possible to conclude in
general on the modifying impact of operators on alternatives.
\end{enumerate}

\end{document}